\documentclass{article}

\PassOptionsToPackage{numbers, compress}{natbib}
\usepackage[final,main,nonanonymous]{neurips_2026}
\usepackage[utf8]{inputenc}
\usepackage[T1]{fontenc}
\ifdefined\pdfobjcompresslevel
\fi
\usepackage{hyperref}
\usepackage{url}
\usepackage{graphicx}
\usepackage{float}
\usepackage{booktabs}
\usepackage{amsmath}
\usepackage{amsfonts}
\usepackage{nicefrac}
\usepackage{microtype}
\usepackage[table]{xcolor}
\IfFileExists{axessibility.sty}{\usepackage[accsupp]{axessibility}}{}
\usepackage{xspace}
\usepackage{multirow}
\usepackage{cleveref}
\usepackage{bm}

\newcommand{\eg}{\emph{e.g.}\xspace}
\newcommand{\ie}{\emph{i.e.}\xspace}

\newcommand{\method}{CoLa3D\xspace}

\newcommand{\heading}[1]{\vspace{.5em}\noindent\textbf{#1.}}
\definecolor{besthighlight}{RGB}{255,205,205}
\definecolor{secondhighlight}{RGB}{255,235,80}
\definecolor{thirdhighlight}{RGB}{218,241,214}
\newcommand{\bestcell}[1]{\cellcolor{besthighlight}\textbf{#1}}
\newcommand{\secondcell}[1]{\cellcolor{secondhighlight}\underline{#1}}
\newcommand{\thirdcell}[1]{\cellcolor{thirdhighlight}\textit{#1}}
\newcommand{\besthl}[1]{{\setlength{\fboxsep}{1pt}\colorbox{besthighlight}{\textbf{#1}}}}
\newcommand{\secondhl}[1]{{\setlength{\fboxsep}{1pt}\colorbox{secondhighlight}{\underline{#1}}}}
\newcommand{\thirdhl}[1]{{\setlength{\fboxsep}{1pt}\colorbox{thirdhighlight}{\textit{#1}}}}

\newcommand{\priormask}{\raisebox{-0.25\height}{\includegraphics[height=2.2ex]{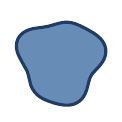}}}
\newcommand{\priorpoint}{\raisebox{-0.25\height}{\includegraphics[height=2.2ex]{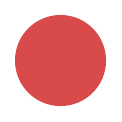}}}
\newcommand{\priornone}{\raisebox{-0.25\height}{\includegraphics[height=2.2ex]{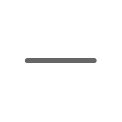}}}

\title{\method: Composable Latent 3D Decomposition}
\author{Xudong Cai$^{1,3}$ \quad Yongcai Wang$^{1,2*}$ \quad Deying Li$^{1}$ \quad Chuanxia Zheng$^{3*\dagger}$ \\ [0.5em]
  $^{1}$School of Information, Renmin University of China \\
  $^{2}$Hebei Key Laboratory of Real-virtual Integrated Autonomous Systems (RIAS) \\
  $^{3}$PVG, Nanyang Technological University \\ [0.3em]
  \url{https://master-cai.github.io/CoLa3D/}
}

\begin{document}
\maketitle
\begingroup
\renewcommand{\thefootnote}{\fnsymbol{footnote}}
\footnotetext[1]{Corresponding authors.\quad $^{\dagger}$Project lead.}
\endgroup

\begin{figure}[!ht]
    \centering
    \vspace{-0.6em}
    \includegraphics[width=\linewidth]{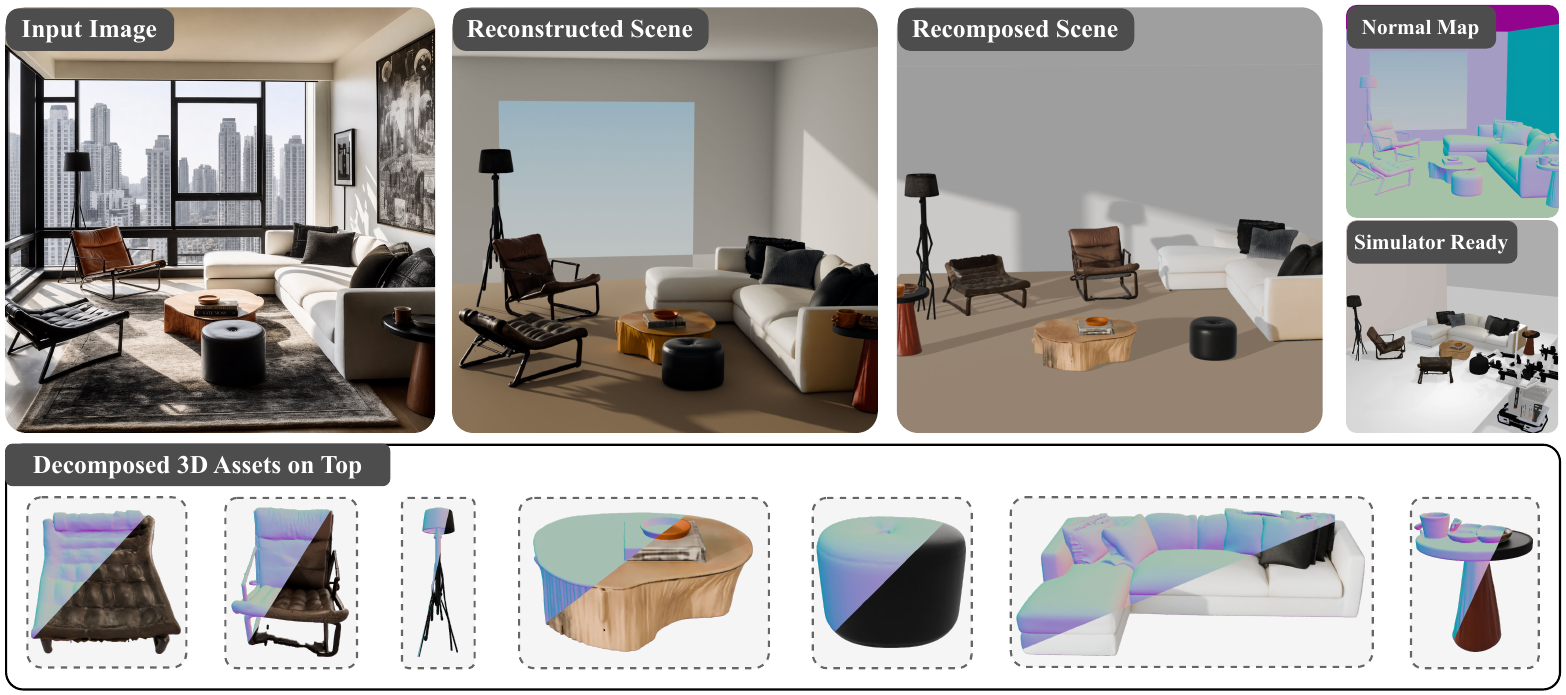}
    \vspace{-1em}
    \caption{\textbf{Composable 3D scene generated by \method.}
    Given a single input image (\emph{left}),
    \method decomposes its \emph{whole}-scene latent into individual object meshes (\emph{bottom}) within a shared coordinate frame,
    faithfully reproducing the input layout (\emph{middle left}) while keeping each object physically accessible for downstream tasks such as rearrangement (\emph{middle right}) and physical simulation (\emph{right}).}
    \label{fig:teaser}
\end{figure}

\begin{abstract}

We introduce \method,
a simple and effective model for composable 3D scene decomposition that,
given a single \emph{whole}-scene latent from either a single image or an existing scene mesh as input,
directly outputs a set of object meshes with consistent spatial layout
that can be assembled back into a coherent scene.
\method introduces a novel paradigm for 3D scene decomposition within a shared \emph{whole}-scene latent,
unlike alternatives that perform per-object generation in canonical spaces and then glue the objects back together with fragile pose and scale estimation.
The network is a lightweight, flexible, and scalable query-based transformer that accepts various types of queries,
including 2D masks, 3D points, and learnable queries,
and learns to decompose the scene into objects in a compact latent space.
On both synthetic and real-world datasets, \method achieves state-of-the-art performance on composable 3D scene decomposition, substantially improving global scene-level consistency and object-level quality while running $2\times$ faster and using less than $1\%$ of SAM 3D's training data.
Visible depth on par with MoGe2 further confirms that latent decomposition preserves accurate visible scene geometry while producing physically complete, decomposable object meshes.

\end{abstract}

\section{Introduction}
\label{sec:intro}

We consider the problem of \emph{composable 3D scene decomposition} from either an existing 3D mesh or a single image.
The output represents the scene as a coherent whole and a collection of \emph{physically plausible 3D instances},
namely object-level meshes together with their spatial arrangement (\cref{fig:teaser}).
This is the form in which our physical world is organized:
a living room,
for example,
is not merely a pile of \emph{whole} scene geometry,
but a structured arrangement of sofas, coffee tables, lamps, and other objects.
Recovering such a composable 3D scene is important for embodied agents that reason jointly about objects and their relationships to interact with them in the physical world~\cite{li2023behavior,IchterBCFHHHIIJ22,ChernyadevB0LSJ24},
as well as for AR/VR systems that require editable and interactive scene representations~\cite{niemeyer2021giraffe,wu2022object,yang2023urbangiraffe}.

Traditional 3D decomposition,
also known as \emph{instance segmentation},
operates on raw 3D observations such as point clouds~\cite{qi2017pointnet,zhao2021point,liu2023partslip,mask3d,zhou2025pointsam,pmlr-v267-mcvay25a},
but often struggles when the geometry is noisy or incomplete.
Another line of work reconstructs composable 3D scenes from images or RGB-D scans,
but typically relies on retrieval-based pipelines~\cite{Avetisyan_2019_CVPR,kuo2020mask2cad,huang2025litereality} or per-scene optimization~\cite{Zhi_2021_ICCV,Yang_2021_ICCV,wu2022object,wu2024clusteringsdf}.
These methods are difficult to scale in practice due to the limited coverage of retrieval libraries and the computational cost of test-time optimization.
These methods either falter on noisy geometry or rely on costly external pipelines, motivating learning-based composable reconstruction from images.

Recent progress in 3D object generators~\cite{zhang20233dshape2vecset,zhang2024clay,xiang2024structured,TripoSR2024,li2025triposg,hunyuan3d22025tencent,xiang2025native} and 2D segmenters~\cite{kirillov2023segment,ravi2024sam2,carion2026sam3} has made composable 3D scene reconstruction from sparse views, or even a single image, increasingly plausible.
Recent contributions like MIDI~\cite{huang2025midi}, CAST~\cite{yao2025cast}, SceneGen~\cite{meng2026scenegen}, and SAM 3D~\cite{chen2026sam} follow a common recipe: they first segment objects in 2D image space using SAM~\cite{kirillov2023segment}, and then condition 3D generators on these masks to reconstruct object assets.
While promising,
the recomposed scene often suffers from spatial inconsistency and incoherence, as illustrated in~\cref{fig:vis_synthetic} and~\cref{fig:vis_real}.
The core difficulty is that object geometry and scene-level scale, position, and pose are not jointly modeled in a shared 3D coordinate frame:
object generators usually operate in \emph{canonical} spaces, and the relative layout has to be glued back across separately-generated object latents.
Very recent methods~\cite{huang2025midi,meng2026scenegen,chen2026sam,lin2025partcrafter} mitigate this issue by introducing cross-attention between object latents or by conditioning on global scene features,
but they still couple objects only indirectly rather than decomposing a scene into object representations in the same coordinate frame.
Moreover, these approaches must generate multiple object latents via diffusion models,
making them computationally slow.

In this paper, we introduce a \emph{novel paradigm} for joint object decomposition within a shared scene latent, instead of independent per-object generation in canonical spaces.
We do so by introducing \emph{\textbf{Co}mposable \textbf{La}tent \textbf{3D} decomposition} (\textbf{\method}),
a feed-forward neural network that predicts a set of \emph{object} latents from a single \emph{whole-scene} latent in the scene coordinate frame,
which are then decoded in \emph{parallel} to produce \emph{individual} meshes,
as well as their recomposed \emph{whole} scene mesh.
Because its decomposition is performed in a \emph{shared} scene coordinate frame, the decoded individual meshes are naturally more spatially consistent with the global scene geometry.
Remarkably,
its outputs exhibit less interpenetration and appear more physically plausible than MIDI, SceneGen, PartCrafter, and SAM 3D,
making them readily usable in physical simulators for downstream tasks.

We instantiate this idea with a flexible query-based decomposition transformer that performs all object decomposition in a single forward pass.
Concretely,
\method accepts as input (i) the whole-scene latent, and (ii) promptable queries (\eg, 2D masks, 3D points, or purely learnable queries) that interact with the latent tokens,
and produces a set of object latents for a generative decoder to produce per-object meshes.
To further improve generalization to novel categories and scenes,
we incorporate 3D semantic features from Utonia~\cite{zhang2026utonia} and 2D semantic features from DINOv3~\cite{simeoni2025dinov3}.

To evaluate our model,
we create challenging benchmarks on both synthetic (3D-FRONT~\cite{fu20213d} and MetaScenes~\cite{yu2025metascenes}) and real-world datasets (Replica~\cite{replica19arxiv} and ScanNet++~\cite{yeshwanth23scannet:}).
\method achieves state-of-the-art performance on both object asset quality and global scene consistency.
More impressively, compared with SAM 3D~\cite{chen2026sam}, which was trained on a large-scale dataset with SAM-based priors and substantial policy tuning and computational resources,
\method shows better results using less than $1\%$ of SAM 3D's training data, demonstrating the effectiveness of our latent decomposition paradigm.
We hope this work will inspire future research on composable 3D scene generation.

Recent promptable 3D segmenters like Point-SAM~\cite{zhou2025pointsam}, Locate 3D~\cite{pmlr-v267-mcvay25a}, and Particulate~\cite{li2025particulate} also build on query-based architectures, but they operate on raw 3D observations.
In contrast, \method operates on a compact generative 3D latent.
This allows \method to extract object-level meshes, rather than segmentation masks, from encoded meshes or single-image generated scene latents.

To summarize, we make the following contributions:
(1) We introduce a novel paradigm, \method, for composable 3D scene decomposition: it obtains object latents from one whole-scene latent in a shared scene coordinate frame and decodes them in parallel into per-object meshes.
(2) We design a lightweight query-based decomposition transformer that performs latent-space object extraction in a single forward pass, using 2D mask, 3D point, and learnable queries.
(3) We show that \method achieves state-of-the-art performance on both synthetic and real-world datasets, outperforming strong baselines such as MIDI~\cite{huang2025midi}, SceneGen~\cite{meng2026scenegen}, PartCrafter~\cite{lin2025partcrafter}, and SAM 3D~\cite{chen2026sam} in both object quality and global scene consistency.

\section{Related Work}\label{sec:related}

\heading{3D Reconstruction and Generation}
Recovering 3D geometry from images is a long-standing problem in computer vision.
Classical reconstruction pipelines such as COLMAP~\cite{schonberger2016sfm,schonberger2016mvs} chain multiple stages---structure-from-motion and multi-view stereo---to estimate camera poses and dense geometry, whereas recent feed-forward models~\cite{wang2024dust3r,chen2024mvsplat,wang2026pi,wang2025vggt,wang2025continuous,lin2025depth,wang2026vggt} unify these stages into an end-to-end network.
Beyond reconstruction, generative 3D models synthesize assets from learned priors with point, voxel, implicit, or Gaussian representations~\cite{nichol2022point,ren2024xcube,park2019deepsdf,mildenhall2020nerf,sitzmann2021light,kerbl20233d}, and recent foundation 3D generators~\cite{zhang2024clay,xiang2024structured,xiang2025native,yang2024hunyuan3d,hunyuan3d22025tencent,hunyuan3d21,li2025triposg} further produce high-quality assets in compact 3D latent spaces~\cite{zhang20233dshape2vecset}.
However, these holistic generators do not expose editable instance-level structure.
Part-aware 3D generation~\cite{liu2024part123,chen2024partgen,gao2025meshart,omnipart,partpacker} addresses object decomposition but focuses on single objects rather than scene-level instances.
We instead target \emph{scene}-level 3D instance decomposition.

\heading{Scene Decomposition}
Scene decomposition aims to obtain a composable 3D scene made of individual object instances.
Traditional 3D decomposition performs instance segmentation on raw 3D observations~\cite{mask3d,oneformer3d,zhou2025pointsam,pmlr-v267-mcvay25a,li2025particulate}, but they are sensitive to noisy observations and output masks instead of complete object meshes.
Recently, generative scene decomposition instead targets object-level geometry together with a coherent global layout.
Staged pipelines first segment scene components and then retrieve, reconstruct, or optimize them separately~\cite{kuo2020mask2cad,roca,diffcad,Zhi_2021_ICCV,Yang_2021_ICCV,wu2022object,wu2024clusteringsdf,ni2025dprecon,zhu2025objectgs,gen3dsr,reparo,yao2025cast,zhao2025depr}.
They are modular but often depend on external segmentation, per-object generation, and layout refinement, which can lead to inconsistent pose, scale, and contact.
Joint models couple object geometry, pose, and instance assignment during generation~\cite{coherent3d,huang2025midi,moca,wang2026scenetransporter,meng2026scenegen,chen2026sam,threedfix,iscene}, yet objects are still generated in separate latents and need extra pose and scale estimation.
PartCrafter~\cite{lin2025partcrafter} jointly denoises per-object latent in a global frame, but object poses are still implicitly modeled through attention.
In contrast, our method decomposes a single whole-scene latent into globally aligned object latents in a feed-forward manner, avoiding separate object generation and per-scene optimization. Inter-object spatial relations are thus naturally encoded in the shared latent.

\heading{Query-Based Models}
Query-based models predict object sets through learnable queries and have become a standard formulation for detection and segmentation~\cite{carion2020end,deformable_detr,dino,cheng2022masked,maskdino}.
Recent promptable 3D models also use points or masks as queries for 3D segmentation~\cite{zhou2025pointsam,ma2026p,zhu2026partsam}.
Related object-centric models also learn latent slots to represent scene entities~\cite{slot_attention,slate,glass}.
Our method uses queries as latent extractors: each query interacts with a compact foundation-model scene latent and outputs an object latent that can be decoded into a mesh within the shared scene coordinate frame.

\section{Method}
\label{sec:method}

\begin{figure*}[t]
    \centering
    \includegraphics[width=\textwidth]{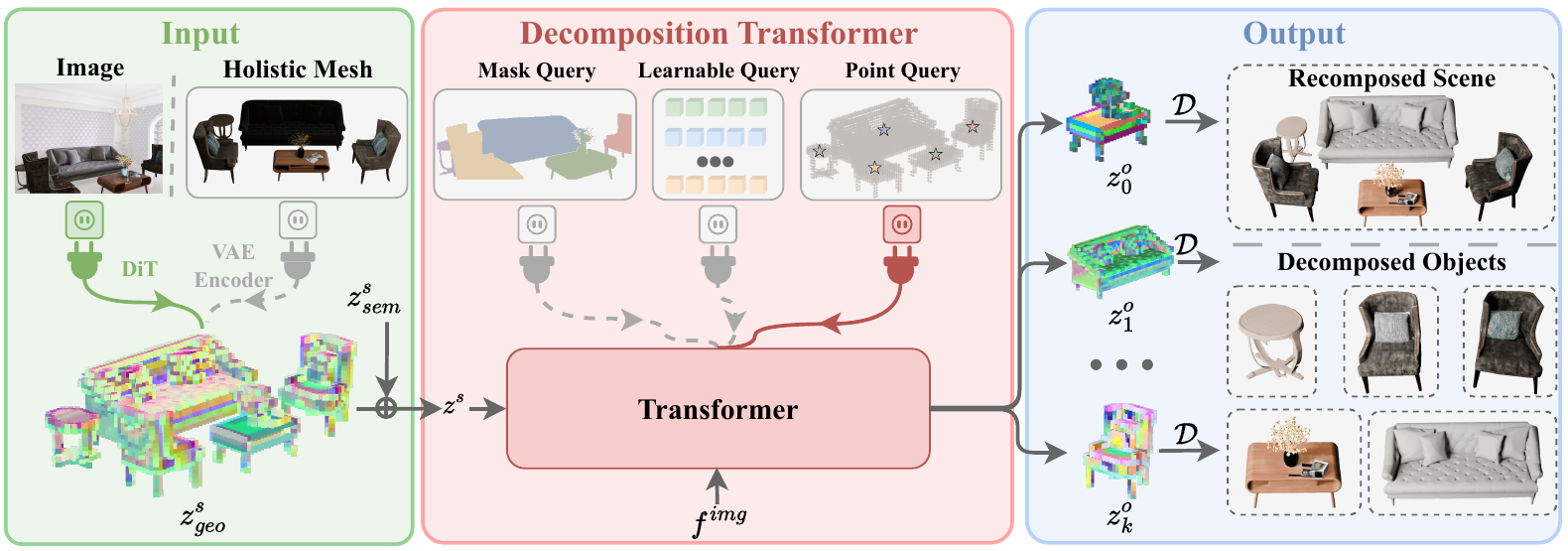}
    \vspace{-1em}

    \caption{\textbf{Overview of \method.} Starting from an RGB image or a holistic scene mesh, \method first obtains a structured scene latent $\bm{z}^s_\text{geo}$ with TRELLIS.2~\cite{xiang2025native} and augments it with 3D semantic features $\bm{z}^s_\text{sem}$ from Utonia~\cite{zhang2026utonia}.
    A lightweight decomposition transformer then uses mask, learnable, or point queries to decompose $\bm{z}^s = \bm{z}^s_\text{geo} + \bm{z}^s_\text{sem}$ through $L$ blocks of self-attention, cross-attention to frozen DINOv3~\cite{simeoni2025dinov3} features $\bm{f}^\text{img}$, and alternating Latent-to-Query/Query-to-Latent cross-attention.
    Finally, a multi-hot assignment extracts per-instance latents $\{\bm{z}^o_k\}$, which are decoded in parallel by the shared decoder $\mathcal{D}$ into aligned object meshes and can be recomposed in the global 3D space.
    }
    \label{fig:pipeline}

\end{figure*}

Given a single image $\bm{I}^s$ or an existing 3D scene mesh
$\mathcal{M}^s$
from AI-generated, scanned, or artist-created sources,
our goal is to obtain a composable 3D scene as a set of object meshes
$\{\hat{\mathcal{M}}^o_k\}_{k=1}^{K}$
with their correct spatial layout,
where $K$ is the number of objects in the scene.
Here, we introduce \method,
a feed-forward neural network that
predicts a set of object latents from a single whole-scene latent in a \emph{global latent space},
which are then decoded in parallel to produce per-object meshes.
The designed query-based decomposition transformer is efficient, flexible, and scalable.
It accepts various types of queries and operates directly on the compact 3D scene latent (as illustrated in \Cref{fig:pipeline}),
instead of segmenting the scene in 2D and lifting each instance separately as in prior methods~\cite{huang2025midi,meng2026scenegen,chen2026sam},
or performing segmentation on raw 3D observations~\cite{mask3d,oneformer3d,zhou2025pointsam,pmlr-v267-mcvay25a,li2025particulate}.

\subsection{Preliminaries}
\label{sec:foundational_model}
\method is built upon TRELLIS.2~\cite{xiang2025native},
a state-of-the-art image-to-3D generator with a compact structured 3D latent space.
The proposed latent decomposition paradigm is not specific to TRELLIS.2
and should be applicable to other foundational 3D generators,
such as Hunyuan3D~\cite{yang2024hunyuan3d,hunyuan3d22025tencent}.

Given a 3D mesh $\mathcal{M}$,
it is first represented by O-Voxel
$\mathcal{O}(\mathcal{M})\mapsto\{\bm{f}_i, \bm{p}_i\}_{i=1}^{L_\text{vox}}$
in a regular 3D grid
$\Omega\in\{0, 1, \cdots, N-1\}^3$,
where $\bm{f}_i$ is the feature of the $i$-th active cell and $\bm{p}_i$ is its 3D coordinate,
and $L_\text{vox}$ is the number of active cells.
Then,
the O-Voxel is encoded into a compact latent $\bm{z} = \mathcal{E}(\mathcal{O}(\mathcal{M}))$ by a sparse-compression encoder,
which can be decoded back to a mesh $\hat{\mathcal{M}} = \mathcal{D}(\bm{z})$ by a corresponding sparse-convolution decoder.
Here,
$\bm{z} = \{\hat{\bm{f}}, \hat{\bm{p}}\}_{i=1}^{L_\text{lat}}$,
where
$\hat{\bm{f}}$ and $\hat{\bm{p}}$ are the updated feature and coordinate of the $i$-th active cell in the latent space,
and $L_\text{lat} \ll L_\text{vox}$ is the active cell number in the latent space,
achieving a highly compact representation of the original mesh.
During image-to-3D inference,
TRELLIS.2 uses image-conditioned flow models~\cite{lipman2023flow,peebles2023scalable}
to predict the sparse structure and then generate geometry latents.

We accordingly obtain the scene geometry latent $\bm{z}^s_\text{geo}$ either by encoding $\mathcal{M}^s$ with the O-Voxel encoder, or by generating it from $\bm{I}^s$ with the image-conditioned flow models.
However,
TRELLIS.2 is originally designed for object-level generation,
and thus we need to adapt it for scene-level generation and decomposition.
Since scene-level fine-tuning is not our main contribution,
we describe the details in~\cref{sec:appendix_training} and focus on the novel latent decomposition module in the below.

\subsection{Query-Based 3D Latent Decomposition Transformer}
\label{sec:ladm}
The input is a composite whole-scene latent
$\bm{z}^s = \bm{z}^s_\text{geo} + \bm{z}^s_\text{sem}$,
where the geometry component $\bm{z}^s_\text{geo}$ is produced by a foundational 3D generator TRELLIS.2~\cite{xiang2025native},
and the semantic component $\bm{z}^s_\text{sem}$ by a 3D semantic foundation model Utonia~\cite{zhang2026utonia}; see \textit{Feature Enhancement} below for details.
\method's latent transformer maps this scene latent to a set of object latents:
\begin{equation}
    \mathcal{Z}= f_\theta(\bm{z}^s, \mathcal{Q}),
\end{equation}
where
$\mathcal{Z} = \{\bm{z}^o_k\}_{k=1}^{K}$,
$\bm{z}^o_k$ is the latent of the $k$-th object,
which can be decoded into an object mesh $\hat{\mathcal{M}}^o_k$ with a shared decoder,
and $\mathcal{Q}$ is a set of queries that can be 2D masks, 3D points, or purely learnable queries.
Unlike recent query-based 3D segmentation methods~\cite{zhou2025pointsam,li2025particulate,pmlr-v267-mcvay25a} that operate on raw point clouds or self-learned 3D features,
our input $\bm{z}^s$ \emph{carries richer geometric and semantic priors while remaining far more compact than raw 3D observations}.

\heading{Query Encoder}
One key ingredient is the set of multi-modal queries $\mathcal{Q}$, which specify objects to be extracted from the scene latent $\bm{z}^s$.
In this work, we consider three query types: 2D masks ($\mathcal{Q}^{\mathrm{M}}$), 3D points ($\mathcal{Q}^{\mathrm{P}}$), and learnable queries ($\mathcal{Q}^{\mathrm{L}}$).
Each query produces a $d$-dimensional token $\bm{q}_k$ via a shared two-step recipe (encode $\rightarrow$ MLP); only the encode step is type-specific.
For brevity, $K$ denotes the active query count, which equals $K_{\max}$ in the learnable mode.

For \textbf{mask prompts} $\{\mathbf{M}_k\}_{k=1}^K$, where $\mathbf{M}_k\in\{0,1\}^{H\times W}$ is the binary image mask of object $k$, we first use a lightweight CNN with global average pooling to extract a feature vector $\bm{f}_k^\text{mask}\in\mathbb{R}^d$.
To disambiguate masks of similar shape, we additionally encode the spatial location of each mask,
by applying sinusoidal positional encoding~\cite{vaswani2017attention} to the center location of the mask,
and map the result to a vector $\bm{f}_k^\text{pos}\in\mathbb{R}^d$ via an MLP.
The final query token is then given by
$\bm{q}_k = \operatorname{MLP}(\operatorname{LayerNorm}(\bm{f}_k^\text{mask} + \bm{f}_k^\text{pos}))$,
yielding the mask query set $\mathcal{Q}^{\mathrm{M}}=\{\bm{q}_k\}_{k=1}^K$.

\textbf{Point prompts} $\{\mathbf{P}_k\}_{k=1}^K$, where $\mathbf{P}_k\in\mathbb{R}^3$ is the 3D coordinate of a prompted active latent cell, are encoded with positional encoding and projected to $\bm{f}_k^\text{point}\in\mathbb{R}^d$ with an MLP.
We also augment it with the corresponding latent features at the prompted point,
which is then mapped to obtain $\bm{f}_k^\text{feat}\in\mathbb{R}^d$ using another MLP.
The query token $\bm{q}_k$ is finally obtained by adding them together, yielding the point query set $\mathcal{Q}^{\mathrm{P}}=\{\bm{q}_k\}_{k=1}^K$.

We also consider \textbf{learnable queries},
without using any explicit priors.
Since the number of objects in the scene is unknown in advance,
we maintain a learnable query set $\mathcal{Q}^{\mathrm{L}}=\{\bm{q}_k\}_{k=1}^{K_{\max}}$,
with $K_{\max}$ set larger than the typical object count.
Each slot $k$ is assigned a normalized index $k/(K_{\max}-1)\in[0,1]$, encoded with sinusoidal positional encoding, and mapped to $\bm{q}_k$ through a 2-layer MLP.

\heading{Feature Enhancement}
To improve the generalization to novel categories and scenes,
we augment the geometry latent
$\bm{z}^s_\text{geo}$ with 3D semantic features
$\bm{z}^s_\text{sem}$
obtained from Utonia~\cite{zhang2026utonia}.
Specifically,
we use the input mesh when $\mathcal{M}^s$ is provided;
for image input, we first decode the generated scene latent into a proxy mesh.
The mesh is then mapped to an O-Voxel representation
$\{\bm{f}_i, \bm{p}_i\}_{i=1}^{L_\text{vox}}$,
in a regular 3D grid
$\Omega\in\{0, 1, \cdots, N-1\}^3$.
Then,
we feed the
$\{\bm{p}_i\}_{i=1}^{L_\text{vox}}$,
where $\bm{p}_i\in\mathbb{R}^3$ is the 3D position of the $i$-th active cell,
into Utonia to obtain the corresponding semantic features $\{\bm{f}_i^\text{sem}\}_{i=1}^{L_\text{lat}}$.
To align the dimensions of the geometry and semantic features,
we apply an MLP to map
$\{\bm{f}_i^\text{sem}\}_{i=1}^{L_\text{lat}}$
to
$\bm{z}^s_\text{sem} = \{\hat{\bm{f}}_i^\text{sem}\}_{i=1}^{L_\text{lat}}$.
After this operation,
we obtain scale and dimensionality aligned geometry features $\bm{z}^s_\text{geo} \in \mathbb{R}^{L_\text{lat} \times d}$
and semantic features $\bm{z}^s_\text{sem} \in \mathbb{R}^{L_\text{lat} \times d}$,
where $L_\text{lat}$ is the number of latent active cells.
The enhanced scene latent is then obtained by summing them together
$\bm{z}^s = \bm{z}^s_\text{geo} + \bm{z}^s_\text{sem}.$

\heading{Attention Blocks}
Following recent query-based segmentors~\cite{kirillov2023segment,li2025particulate,pmlr-v267-mcvay25a,li2026instruct},
we apply a stack of attention blocks to let queries interact with the scene latent and specialize into instance-level representations.
In each block,
we first perform self-attention over query tokens
$\{\bm{q}_k\}_{k=1}^K$.
Then,
updated query tokens
$\{\hat{\bm{q}}_k\}_{k=1}^K$
and latent tokens
$\{\bm{z}^s_i\}_{i=1}^{L_\text{lat}}$
cross-attend to the DINOv3~\cite{simeoni2025dinov3} features $\bm{f}^\text{img}$, grounding the 3D representations with 2D semantic cues.
Finally,
alternating Latent-to-Query and Query-to-Latent cross-attention are performed to flow information between them.
We apply 3D RoPE~\cite{su2024roformer,wang2024qwen2vl} to the query self-attention and bidirectional query--latent cross-attention layers, preserving relative 3D positional cues.
After several blocks,
the query and latent tokens are refined into instance-aware representations.
Note that,
the number of queries $K$ is significantly smaller than the number of latent cells $L_\text{lat}$ and DINOv3 features $h\times w$,
so the query self-attention and other cross-attention operations are efficient and scalable to complex scenes with many objects and high-resolution latents.
See~\cref{sec:appendix_attention} for block diagram and positional encoding details.

\heading{Instance Prediction}
After obtaining the refined query and latent tokens
$\{\hat{\bm{q}}_k\}_{k=1}^K$ and $\{\hat{\bm{z}}^s_i\}_{i=1}^{L_\text{lat}}$,
we predict a soft assignment matrix
$\mathbf{S}\in[0,1]^{L_\text{lat}\times K}$,
whose $(i,k)$-entry $s_{i,k}$ is the probability of latent cell $i$ belonging to object query $k$.
Specifically,
we first project tokens into a common feature space with a shared lightweight MLP,
and then compute $s_{i,k}$ by a temperature-scaled cosine similarity passed through a sigmoid:
\begin{equation}
    s_{i,k} = \operatorname{Sigmoid}\!\left(\tau \cdot \frac{(\bm{z}^s_i)^{\top}\hat{\bm{q}}_k}{\|\bm{z}^s_i\|_2\,\|\hat{\bm{q}}_k\|_2}\right),
\end{equation}
where $\tau$ is a learnable temperature.
For a discrete mask,
we threshold $s_{i,k}$ at $0.5$ into a binary assignment $m_{i,k}\in\{0,1\}$,
with an argmax fallback for cells left unclaimed.
This design allows a latent cell to be assigned to multiple queries since the latent grid is compact and coarse,
where neighboring objects may share cells at contact regions.
More inference details are provided in~\cref{sec:appendix_inference}.

\subsection{Training}
\label{sec:training}

\heading{Training Losses}
We train \method's latent transformer with a loss: $\mathcal{L}=\mathcal{L}_{\mathrm{mask}}+\lambda_{\mathrm{reg}}\,\mathcal{L}_{\mathrm{reg}}$,
where $\mathcal{L}_{\mathrm{mask}}$ is a mask loss that supervises the predicted masks,
and $\mathcal{L}_{\mathrm{reg}}$ is a regularization term that encourages query diversity to prevent query collapse.

Differently from~\cite{pmlr-v267-mcvay25a,li2025particulate},
our mask is supervised in the latent space with a multi-hot assignment (data processing is described in~\cref{sec:appendix_dataset}).
To calculate the mask loss,
we need to match the predicted labels to the ground-truth objects.
For mask and point prompts,
we directly use the prompt-object correspondence to assign labels.
For learnable queries,
following~\cite{carion2020end,li2025particulate}, we apply Hungarian matching with a mask-to-mask cost,
since the queries are unordered.
In both cases we obtain an assignment $\pi(k)$ that maps each ground-truth object $k$ to its corresponding query index.
Then, the mask loss averages the BCE and Dice terms over all matched objects:
\begin{equation}
    \mathcal{L}_{\mathrm{mask}}=
    \frac{1}{|\mathcal{K}|}\sum_{k\in\mathcal{K}}\!\Big[
        \lambda_{\mathrm{bce}}\,\mathrm{BCE}\!\left(\mathbf{S}_{:,\pi(k)},\,\bm{y}^{*}_{k}\right)
        + \lambda_{\mathrm{dice}}\,\mathrm{Dice}\!\left(\mathbf{S}_{:,\pi(k)},\,\bm{y}^{*}_{k}\right)
        \Big],
\end{equation}
where $\mathbf{S}_{:,\pi(k)}\in[0,1]^{L_\text{lat}}$ is the column of $\mathbf{S}$ for the query matched to object $k$,
$\bm{y}^{*}_{k}\in\{0,1\}^{L_\text{lat}}$ is the multi-hot ground-truth mask of object $k$ over the $L_\text{lat}$ active latent cells,
$\mathcal{K}=\{k:\bm{y}^{*}_{k}\neq\bm{0}\}$ collects the objects with non-empty masks,
and $\lambda_{\mathrm{bce}}$, $\lambda_{\mathrm{dice}}$ balance the two terms.
The BCE term provides dense cell-wise supervision,
while the Dice term (formulated in~\cref{sec:appendix_training}) normalizes each matched object mask and reduces the bias of cell-wise BCE toward large objects.

To prevent \emph{query collapse},
where multiple queries specialize to the same object and leave others unused,
we apply a regularizer $\mathcal{L}_{\mathrm{reg}}$ penalizing pairwise cosine similarity among refined queries.
Denoting by
$\tilde{\bm{q}}_{k}=\hat{\bm{q}}_{k}/\|\hat{\bm{q}}_{k}\|_{2}$
the $\ell_{2}$-normalized query tokens,
the regularizer is defined as
\begin{equation}
    \mathcal{L}_{\mathrm{reg}}
    =\frac{1}{K_{\max}(K_{\max}-1)}
        \sum_{i=1}^{K_{\max}}\sum_{\substack{j=1;j\neq i}}^{K_{\max}}
        \big|\tilde{\bm{q}}_{i}^{\top}\tilde{\bm{q}}_{j}\big|.
\end{equation}
This spreads queries across the feature space and encourages each to attach to a distinct instance,
which is especially important for learnable queries that have no explicit disambiguating priors.

\heading{Implementation Details}
By default,
\method's latent transformer uses
$K_{\max}{=}8$ for learnable queries,
and employs $L{=}6$ attention blocks with a hidden dimension of $d{=}768$.
The model was trained using 3D-FRONT dataset~\cite{fu20213d},
where we combine the training splits prepared by MIDI~\cite{huang2025midi} and InstPIFU~\cite{liu2022instpifu},
with a total of around 13k scenes and 82k images.
We train the latent transformer for 60k steps with the Adam optimizer, an initial learning rate of $1\times10^{-4}$, and a per-GPU batch size of 16.
The loss weights are set to $\lambda_{\mathrm{bce}}{=}1.0$, $\lambda_{\mathrm{dice}}{=}1.0$, and $\lambda_{\mathrm{reg}}{=}0.1$.
The training takes around 2 days on 4 NVIDIA H100 GPUs.
We also finetune the TRELLIS.2 geometry components on the same training set to adapt it for scene-level generation.
Since this is not the main contribution of this work,
we detail the finetuning process in~\cref{sec:appendix_training}.

\section{Experiments}\label{sec:experiments}

\subsection{Experimental Setup}
\label{sec:exp_setup}

\heading{Datasets}
We assess our model on both synthetic and real-world datasets.
For \emph{synthetic} data,
we use 3D-FRONT~\cite{fu20213d} and MetaScenes~\cite{yu2025metascenes},
which provide complete per-object meshes and layouts.
For 3D-FRONT,
we filter the InstPIFU~\cite{liu2022instpifu} test split and retain 85 high-quality scenes for evaluation;
for MetaScenes,
we analogously select 50 high-quality scenes.
Each synthetic scene is evaluated from one input image.
Furthermore,
to further assess the cross-dataset generalization ability,
we directly evaluate all models on \emph{real-world} datasets,
including Replica~\cite{replica19arxiv} and ScanNet++~\cite{yeshwanth23scannet:}.
For real-world evaluation, we use more challenging scan-derived scenes with reliable ground truth:
8 scenes / 14 images from Replica and 45 scenes / 56 images from ScanNet++.

\heading{Metrics}
For quantitative results,
we normalize predicted and ground-truth point clouds and align them with FilterReg before evaluation,
following~\cite{huang2025midi,meng2026scenegen}.
Then,
we sample 20{,}480 points per object mesh and report Chamfer Distance (CD), F-Score (threshold 0.1), and 3D bounding-box IoU at two levels.
At the \emph{scene level},
we assemble all predicted objects and compare the full scene with the ground truth;
at the \emph{object level}, we compute the same metrics and average over matched objects.
We additionally report per-scene inference time on a single NVIDIA L40S 48 GB GPU, averaging the two datasets in each main table and providing per-dataset timings in Supp.~\cref{tab:supp_timing}.

\heading{Baselines}
We compare with six strong recent single-image scene generation and decomposition baselines:
the per-scene \emph{optimization} method Gen3DSR~\cite{gen3dsr},
and \emph{feed-forward} methods including MIDI~\cite{huang2025midi}, SceneGen~\cite{meng2026scenegen}, PartCrafter~\cite{lin2025partcrafter}, SAM 3D~\cite{chen2026sam}, and 3D-Fixer~\cite{threedfix}.
To further evaluate the merit of our query-based latent decomposition,
we replace our query transformer with Point-SAM~\cite{zhou2025pointsam}:
given the same point prompts as \method (Point), Point-SAM segments the latent cells, and each segmented latent-cell group is decoded by the same object decoder.
For mask-based methods, we provide ground-truth masks as in prior works~\cite{huang2025midi,meng2026scenegen,chen2026sam};
for point-based methods, we randomly sample one point from each ground-truth object.
All baselines use their public \mbox{checkpoints}.

\begin{figure*}[t]
    \centering
    \resizebox{\linewidth}{!}{\includegraphics{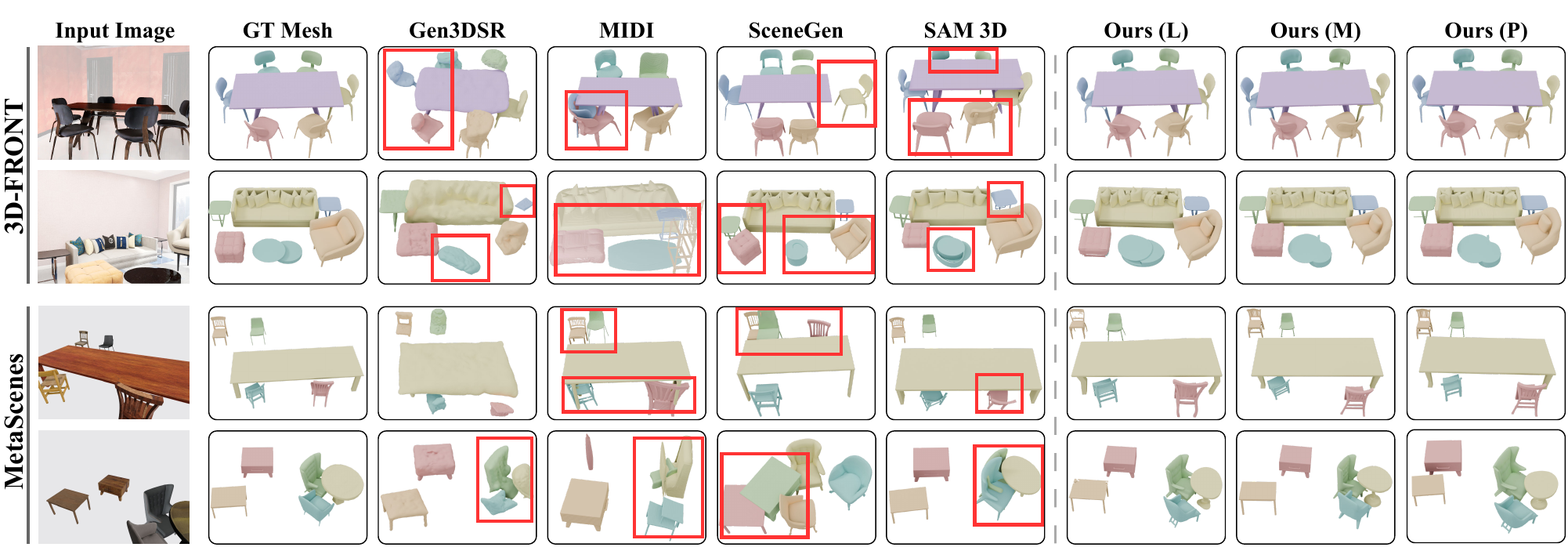}}
    \vspace{-12pt}
    \caption{\textbf{Qualitative comparisons on synthetic data.} Compared with baselines, \method produces spatially coherent object meshes with correct relative positions and scales.}
    \label{fig:vis_synthetic}
\end{figure*}

\begin{table*}[t]
\centering
\caption{\textbf{Comparisons on synthetic datasets.}
\textbf{Prior} column: \protect\priormask{} denotes a 2D instance mask, \protect\priorpoint{} a 3D point prompt, and \protect\priornone{} no prior.
We highlight the \besthl{Best}, \secondhl{Second-Best}, and \thirdhl{Third-Best} results.}
\label{tab:instpifu_results}
\setlength{\tabcolsep}{3pt}
\resizebox{\linewidth}{!}{
\begin{tabular}{@{}l c cccccc cccccc c@{}}
\toprule
\multirow{3}{*}{\textbf{Method}} & \multirow{3}{*}{\textbf{Prior}} & \multicolumn{6}{c}{\textbf{3D-FRONT}~\cite{fu20213d}} & \multicolumn{6}{c}{\textbf{MetaScenes}~\cite{yu2025metascenes}} & \multirow{3}{*}{\textbf{Avg. Time}$\downarrow$} \\
\cmidrule(lr){3-8} \cmidrule(lr){9-14}
& & \multicolumn{3}{c}{\textbf{Scene}} & \multicolumn{3}{c}{\textbf{Object}} & \multicolumn{3}{c}{\textbf{Scene}} & \multicolumn{3}{c}{\textbf{Object}} & \\
\cmidrule(lr){3-5} \cmidrule(lr){6-8} \cmidrule(lr){9-11} \cmidrule(lr){12-14}
& & CD$\downarrow$ & F-Score$\uparrow$ & IoU$\uparrow$ & CD$\downarrow$ & F-Score$\uparrow$ & IoU$\uparrow$ & CD$\downarrow$ & F-Score$\uparrow$ & IoU$\uparrow$ & CD$\downarrow$ & F-Score$\uparrow$ & IoU$\uparrow$ & \\
\midrule
Gen3DSR~\cite{gen3dsr}        & \priormask  & 0.0230 & 0.7625 & 0.5804 & 0.0793 & 0.5211 & 0.4650 & 0.0140 & 0.854 & 0.654 & \secondcell{0.0509} & 0.524 & 0.598 & 229.40 \\
PartCrafter~\cite{lin2025partcrafter} & \priornone  & 0.0468 & 0.6597 & 0.4470 & 0.1626 & 0.4549 & 0.3633 & 0.0460 & 0.6303 & 0.4315 & 0.1291 & 0.4654 & 0.4320 & 51.04 \\
MIDI~\cite{huang2025midi}  & \priormask  & 0.0289 & 0.7014 & 0.4151 & 0.0845 & 0.6512 & 0.5249 & 0.0429 & 0.664 & 0.433 & 0.0697 & 0.617 & 0.522 & 67.45 \\
3D-Fixer~\cite{threedfix}     & \priormask  & 0.0194 & 0.7992 & 0.5455 & 0.0654 & 0.5792 & 0.4517 & 0.0212 & 0.7813 & 0.5581 & 0.0784 & 0.5140 & 0.4486 & 98.35 \\
Point-SAM~\cite{zhou2025pointsam} & \priorpoint & 0.0170 & 0.8811 & 0.7025 & 0.1310 & 0.5892 & 0.4037 & 0.0175 & 0.901 & 0.699 & 0.1315 & 0.546 & 0.402 & 16.87 \\
SceneGen~\cite{meng2026scenegen} & \priormask  & 0.0150 & 0.8434 & 0.6107 & 0.0348 & 0.7582 & 0.6684 & 0.0482 & 0.657 & 0.446 & 0.0593 & 0.630 & 0.543 & 49.30 \\
SAM 3D~\cite{chen2026sam}     & \priormask  & 0.0104 & 0.8889 & 0.6702 & 0.0420 & 0.7143 & 0.5569 & 0.0267 & 0.827 & 0.585 & 0.0794 & 0.612 & 0.440 & 32.15 \\
\midrule
\textbf{\method (Free)}  & \priornone  & \bestcell{0.0075} & \bestcell{0.9263} & \bestcell{0.7474} & \secondcell{0.0300} & \bestcell{0.8252} & \bestcell{0.6933} & \secondcell{0.0069} & \secondcell{0.9252} & \secondcell{0.7566} & \bestcell{0.0425} & \secondcell{0.7782} & \secondcell{0.6632} & \thirdcell{16.64} \\
\textbf{\method (Point)} & \priorpoint & \secondcell{0.0088} & \secondcell{0.9256} & \secondcell{0.7460} & \bestcell{0.0273} & \secondcell{0.8205} & \thirdcell{0.6908} & \bestcell{0.0060} & \bestcell{0.9291} & \bestcell{0.7689} & \thirdcell{0.0510} & \bestcell{0.7996} & \bestcell{0.6982} & \bestcell{16.60} \\
\textbf{\method (Mask)}  & \priormask  & \thirdcell{0.0089} & \thirdcell{0.9166} & \thirdcell{0.7341} & \thirdcell{0.0307} & \thirdcell{0.8144} & \secondcell{0.6927} & \thirdcell{0.0076} & \thirdcell{0.9206} & \thirdcell{0.7474} & 0.0582 & \thirdcell{0.7187} & \thirdcell{0.6189} & \secondcell{16.63} \\
\bottomrule
\end{tabular}
}
\end{table*}

\subsection{Assessing Composable Reconstruction on Synthetic Scenes}
\label{sec:exp_synthetic}

We report results on the 3D-FRONT InstPIFU split and MetaScenes in~\cref{tab:instpifu_results}.
All three \method variants surpass the previous \emph{per-object} generation baselines by a large margin at the scene level,
while achieving strong object-level performance across all three query types.
On 3D-FRONT, \method (Free) reduces scene CD by 27.9\% and improves object F-Score by 8.8\% over the best baseline.
On MetaScenes, \method (Point) reduces scene CD by 57.1\% and improves object F-Score by 26.9\% over the best baseline.
The averaged Time column shows that \method (Free) is also around $2\times$ faster than the feed-forward baselines,
due to its joint decomposition in a global latent space.

The qualitative results in~\cref{fig:vis_synthetic} further illustrate the advantages of our latent decomposition paradigm.
MIDI and Gen3DSR often generate wrong shapes (first, second, and last rows) when the object is heavily occluded,
while SceneGen and SAM 3D produce more accurate shapes but suffer from spatial inconsistency and interpenetration (third and last rows),
especially when the two objects are in close contact (last row).
\method shows no such artifacts and produces more coherent layouts and plausible object scales,
even under heavy occlusion,
suggesting that its joint decomposition in a shared 3D latent space leads to more physically plausible and spatially consistent reconstructions.

\subsection{Assessing Cross-Dataset Generalization on Real-World Scenes}
\label{sec:exp_real}

\begin{figure*}[t]
    \centering
    \includegraphics[width=\linewidth]{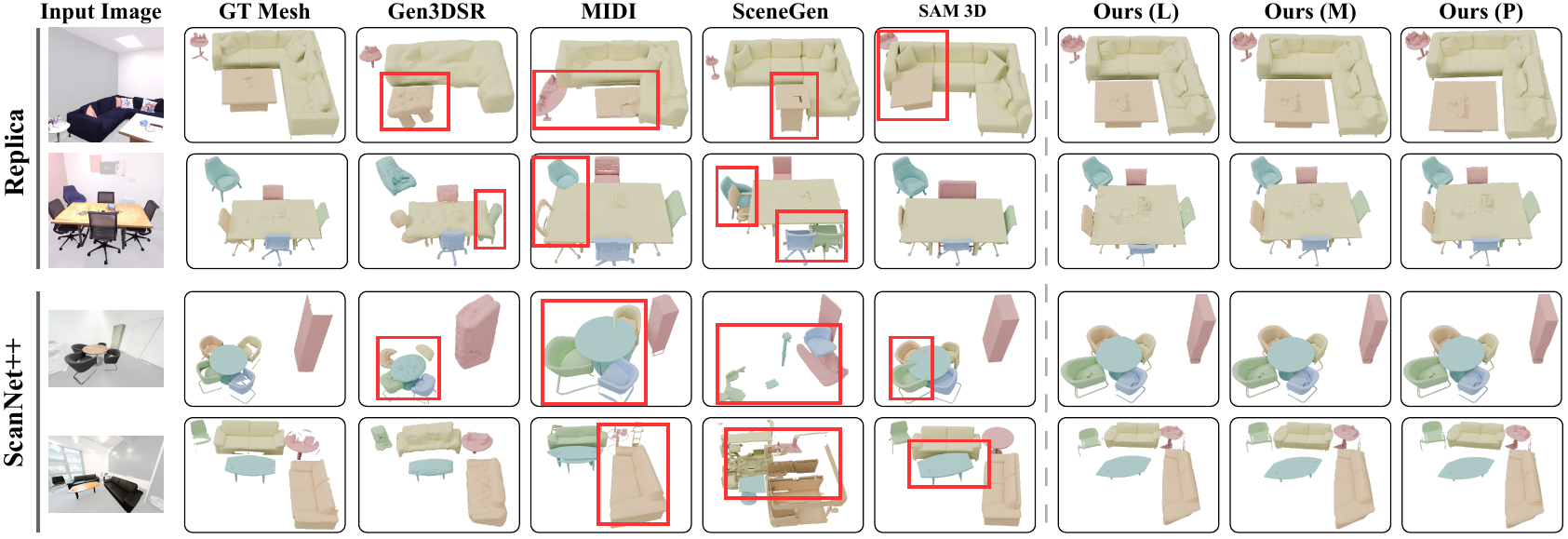}
    \vspace{-12pt}
    \caption{\textbf{Qualitative results on real-world scenes.} \method produces coherent decompositions on Replica and ScanNet++ images, demonstrating robust generalization from synthetic training data.}
    \label{fig:vis_real}
\end{figure*}

\begin{table*}[t]
\centering
\caption{\textbf{Zero-shot generalization on real-world scenes.}
\textbf{Prior} column: \protect\priormask{} denotes a 2D instance mask, \protect\priorpoint{} a 3D point prompt, and \protect\priornone{} no prior.
We highlight the \besthl{Best}, \secondhl{Second-Best}, and \thirdhl{Third-Best} results.
Here, CD is \emph{one-side} Chamfer Distance (GT$\to$Pred).
}
\label{tab:ood_results}
\setlength{\tabcolsep}{3pt}
\resizebox{\linewidth}{!}{
\begin{tabular}{@{}l c cccccc cccccc c@{}}
\toprule
\multirow{3}{*}{\textbf{Method}} & \multirow{3}{*}{\textbf{Prior}} & \multicolumn{6}{c}{\textbf{Replica}} & \multicolumn{6}{c}{\textbf{ScanNet++}} & \multirow{3}{*}{\textbf{Avg. Time}$\downarrow$} \\
\cmidrule(lr){3-8} \cmidrule(lr){9-14}
& & \multicolumn{3}{c}{\textbf{Scene}} & \multicolumn{3}{c}{\textbf{Object}} & \multicolumn{3}{c}{\textbf{Scene}} & \multicolumn{3}{c}{\textbf{Object}} & \\
\cmidrule(lr){3-5} \cmidrule(lr){6-8} \cmidrule(lr){9-11} \cmidrule(lr){12-14}
& & CD$\downarrow$ & Recall$\uparrow$ & IoU$\uparrow$ & CD$\downarrow$ & Recall$\uparrow$ & IoU$\uparrow$ & CD$\downarrow$ & Recall$\uparrow$ & IoU$\uparrow$ & CD$\downarrow$ & Recall$\uparrow$ & IoU$\uparrow$ & \\
\midrule
Gen3DSR~\cite{gen3dsr}           & \priormask  & 0.0070 & 0.837 & 0.612 & \secondcell{0.0244} & 0.558 & 0.576 & 0.0096 & 0.8050 & \thirdcell{0.6444} & \bestcell{0.0333} & 0.4949 & \secondcell{0.4968} & 273.70 \\
PartCrafter~\cite{lin2025partcrafter} & \priornone  & 0.0297 & 0.5623 & 0.3643 & 0.1321 & 0.4070 & 0.3998 & 0.0388 & 0.5912 & 0.4448 & 0.1158 & 0.4482 & 0.3659 & 59.57 \\
MIDI~\cite{huang2025midi}     & \priormask  & 0.0098 & 0.800 & 0.455 & 0.0307 & 0.617 & 0.475 & 0.0197 & 0.6949 & 0.4832 & 0.0468 & 0.5740 & 0.4607 & 78.85 \\
SceneGen~\cite{meng2026scenegen} & \priormask  & 0.0105 & 0.768 & 0.477 & \thirdcell{0.0289} & 0.608 & 0.540 & 0.0280 & 0.6594 & 0.4313 & \thirdcell{0.0417} & 0.5530 & 0.3489 & 61.10 \\
SAM 3D~\cite{chen2026sam}        & \priormask  & 0.0112 & 0.844 & 0.584 & 0.0356 & 0.599 & 0.491 & 0.0100 & 0.8086 & 0.6338 & 0.0450 & 0.6179 & 0.4225 & 37.05 \\
3D-Fixer~\cite{threedfix}        & \priormask  & 0.0069 & \thirdcell{0.8909} & \thirdcell{0.6156} & 0.0333 & 0.5773 & 0.4766 & 0.0207 & 0.7278 & 0.5439 & \secondcell{0.0378} & 0.5471 & 0.4335 & 107.40 \\
\midrule
\textbf{\method (Free)}  & \priornone  & \thirdcell{0.0051} & 0.8898 & 0.5949 & 0.0384 & \thirdcell{0.6575} & \thirdcell{0.5821} & \thirdcell{0.0082} & \thirdcell{0.8310} & 0.6350 & 0.0450 & \bestcell{0.6434} & \bestcell{0.4980} & \secondcell{21.14} \\
\textbf{\method (Point)} & \priorpoint & \bestcell{0.0032} & \bestcell{0.9371} & \secondcell{0.6504} & \bestcell{0.0235} & \bestcell{0.7457} & \bestcell{0.6226} & \secondcell{0.0078} & \secondcell{0.8423} & \bestcell{0.6635} & 0.0470 & \thirdcell{0.6278} & \thirdcell{0.4892} & \thirdcell{21.47} \\
\textbf{\method (Mask)}  & \priormask  & \secondcell{0.0034} & \secondcell{0.9314} & \bestcell{0.6519} & 0.0340 & \secondcell{0.6920} & \secondcell{0.5904} & \bestcell{0.0076} & \bestcell{0.8506} & \secondcell{0.6574} & 0.0456 & \secondcell{0.6419} & 0.4846 & \bestcell{20.83} \\
\bottomrule
\end{tabular}
}
\end{table*}

To further evaluate the generalization ability of \method,
we report results on the real-world Replica and ScanNet++ datasets in~\cref{tab:ood_results}.
Since real datasets provide only partial object meshes,
we report the \emph{one-side} CD (GT$\to$Pred) and Recall in place of bidirectional CD and F-Score.
\Cref{tab:ood_results} shows that \method achieves the best scene-level results on real-world scans and strong object-level performance.
On Replica, \method (Point) reduces scene CD by 53.6\% and improves object Recall by 20.9\% over the best baseline.
On ScanNet++, \method (Mask) reduces scene CD by 20.8\% and \method (Free) improves object Recall by 4.1\% over the best baseline.
\method (Mask) is the fastest according to the averaged Time column.

As evident in~\cref{fig:vis_real}, \method preserves coherent layouts and plausible object scales under zero-shot transfer,
even under heavy occlusion.
In contrast,
Gen3DSR, MIDI, and SceneGen generate wrong shapes (first, third, and last rows).
While SAM 3D is trained on $100\times$ more data than ours and uses SAM-based priors, it still suffers from spatial inconsistency and interpenetration (second and last rows),
suggesting that the per-object paradigm is not friendly to physical plausibility and spatial consistency.
We present qualitative mesh-decomposition results in~\cref{sec:appendix_mesh_decompose}.

\subsection{Assessing Scalability to Denser Scenes}
\label{sec:exp_scalability}

We further evaluate the scalability of \method to denser and more complex scenes by extending both the training data and query capacity.
To this end,
we augment our original training set with ARSG-110K~\cite{threedfix}, a large-scale scene dataset introduced by 3D-Fixer that contains over 110k scenes and 3M annotated images with object-level geometry and layout annotations.
After data cleaning, we retain 100k ARSG scenes and 870k images, with at most 20 objects per scene.
Using the combined training set, we retrain all three \method variants for 120k steps with $K_{\max}{=}20$, keeping the remaining training settings unchanged.

To assess reconstruction quality beyond the original 2--8-object evaluation setting,
we construct \emph{MetaScenes-hard}, a separate test set of 50 MetaScenes scenes containing 9--20 visible objects per scene.
We retain the other filtering criteria used for the main benchmark to ensure reliable reference geometry.
We report CD, F-Score, and bounding-box IoU at both the scene and object levels, measuring the fidelity of the assembled reconstruction and its constituent objects.

\begin{table}[t]
    \centering
    \caption{\textbf{Comparisons on MetaScenes-hard.}
        \textbf{Prior} column: \protect\priormask{} denotes a 2D instance mask, \protect\priorpoint{} a 3D point prompt, and \protect\priornone{} no prior.
        We highlight the \besthl{Best}, \secondhl{Second-Best}, and \thirdhl{Third-Best} results.}
    \label{tab:dense_results}
    \small
    \setlength{\tabcolsep}{6pt}
    \begin{tabular}{@{}l c ccc ccc@{}}
        \toprule
        \multirow{2}{*}{\textbf{Method}} & \multirow{2}{*}{\textbf{Prior}} & \multicolumn{3}{c}{\textbf{Scene}} & \multicolumn{3}{c}{\textbf{Object}}                                                                                         \\
        \cmidrule(lr){3-5} \cmidrule(lr){6-8}
                                         &                                 & CD$\downarrow$                     & F-Score$\uparrow$                   & IoU$\uparrow$       & CD$\downarrow$      & F-Score$\uparrow$   & IoU$\uparrow$       \\
        \midrule
        MIDI~\cite{huang2025midi}        & \priormask                      & 0.0236                             & 0.7235                              & 0.4309              & 0.0766              & 0.5643              & 0.4816              \\
        SceneGen~\cite{meng2026scenegen} & \priormask                      & 0.0414                             & 0.6449                              & 0.4305              & \secondcell{0.0651} & 0.5949              & \thirdcell{0.5249}  \\
        SAM 3D~\cite{chen2026sam}        & \priormask                      & 0.0112                             & 0.8748                              & 0.6339              & 0.0698              & 0.6090              & 0.4668              \\
        3D-Fixer~\cite{threedfix}        & \priormask                      & 0.0184                             & 0.8058                              & 0.5506              & 0.0720              & 0.5238              & 0.4611              \\
        \midrule
        \textbf{\method (Free)}          & \priornone                      & \bestcell{0.0093}                  & \bestcell{0.8921}                   & \bestcell{0.7229}   & 0.0826              & \thirdcell{0.6242}  & 0.5227              \\
        \textbf{\method (Point)}         & \priorpoint                     & \thirdcell{0.0106}                 & \secondcell{0.8910}                 & \thirdcell{0.7112}  & \bestcell{0.0600}   & \bestcell{0.6858}   & \bestcell{0.5745}   \\
        \textbf{\method (Mask)}          & \priormask                      & \secondcell{0.0095}                & \thirdcell{0.8870}                  & \secondcell{0.7180} & \thirdcell{0.0653}  & \secondcell{0.6666} & \secondcell{0.5525} \\
        \bottomrule
    \end{tabular}
\end{table}

\Cref{tab:dense_results} shows that \method retains its scene-level advantage on MetaScenes-hard across all three query types.
Every variant outperforms MIDI, SceneGen, SAM 3D, and 3D-Fixer on scene CD, F-Score, and IoU.
\method (Free) achieves the best scene-level results without instance prompts, reducing CD by 17.0\% over the strongest baseline SAM 3D.
Its scene IoU also increases from 0.6339 to 0.7229 relative to the best baseline.
At the object level, \method (Point) achieves the best CD, F-Score, and IoU.
It reduces CD by 7.8\% over SceneGen and improves F-Score by 12.6\% over SAM 3D, the strongest respective baselines.
These results indicate that \method maintains its scene-level performance advantage over the compared baselines in scenes with more objects.

\subsection{Assessing Visible Depth Accuracy}
\label{sec:exp_depth}

An ideal composable real-to-simulation 3D generation and decomposition should not only produce high-quality object meshes, but also preserve the overall scene geometry and layout.
To evaluate this aspect,
we assess the visible-region depth accuracy of reconstructed meshes on 3D-FRONT.
To this end,
we render 512$\times$512 depth maps from reconstructed meshes and compare them against ground-truth depth.
In addition to the 3D reconstruction baselines, we include MoGe2~\cite{moge2}, a state-of-the-art monocular depth estimator, as a strong reference.
We report scale-and-shift aligned absolute relative error (AbsRel), RMSE, and $\delta_1$ (the fraction of valid pixels with relative depth ratio below 1.25) on visible foreground pixels, together with completion, which measures the fraction of ground-truth foreground pixels covered by valid rendered depth.

\begin{table*}[t]
\centering
\caption{\textbf{Additional results and ablations on 3D-FRONT.} \textbf{Left:} visible-region depth accuracy. MoGe2 is shown as a reference and excluded from ranking. \textbf{Right:} ablation study on the InstPIFU split. We highlight the \besthl{Best}, \secondhl{Second-Best}, and \thirdhl{Third-Best} results among 3D methods.}
\label{tab:depth_ablation_combined}
\small

\begin{minipage}[t]{0.45\textwidth}
\centering
\textbf{(a) Visible-region depth accuracy.}\\[2pt]
\setlength{\tabcolsep}{3pt}
\resizebox{\linewidth}{!}{\begin{tabular}{@{}l c cccc@{}}
\toprule
\textbf{Method} & \textbf{Prior} & AbsRel$\downarrow$ & $\delta_1\uparrow$ & RMSE$\downarrow$ & Completion$\uparrow$ \\
\midrule
\textcolor{gray}{MoGe2~\cite{moge2}} & \textcolor{gray}{--} & \textcolor{gray}{0.0423} & \textcolor{gray}{0.9862} & \textcolor{gray}{0.1802} & \textcolor{gray}{0.9888} \\
\midrule
SAM 3D~\cite{chen2026sam}         & \priormask & 0.0937          & \thirdcell{0.9245} & 0.3444          & 0.8352 \\
SceneGen~\cite{meng2026scenegen}  & \priormask & 0.1282          & 0.8780          & 0.4234          & 0.8494 \\
Gen3DSR~\cite{gen3dsr}            & \priormask & 0.1320          & 0.8636          & 0.4171          & 0.7619 \\
MIDI~\cite{huang2025midi}      & \priormask & 0.1738          & 0.7778          & 0.5447          & 0.8152 \\
\midrule
\textbf{\method (Mask)} & \priormask  & \bestcell{0.0797} & \secondcell{0.9352} & \bestcell{0.3090} & \thirdcell{0.9175} \\
\textbf{\method (Free)} & \priornone  & \secondcell{0.0844} & \bestcell{0.9370} & \secondcell{0.3243} & \secondcell{0.9182} \\
\textbf{\method (Point)} & \priorpoint & \thirdcell{0.0895} & 0.9244          & \thirdcell{0.3251} & \bestcell{0.9337} \\
\bottomrule
\end{tabular}}
\end{minipage}
\hfill
\begin{minipage}[t]{0.51\textwidth}
\centering
\textbf{(b) Ablation study.}\\[2pt]
\setlength{\tabcolsep}{3pt}
\resizebox{\linewidth}{!}{\begin{tabular}{@{}l cccccc@{}}
\toprule
\multirow{2}{*}{\textbf{Model Variant}} & \multicolumn{3}{c}{\textbf{Scene}} & \multicolumn{3}{c}{\textbf{Object}} \\
\cmidrule(lr){2-4} \cmidrule(lr){5-7}
& CD$\downarrow$ & F-Score$\uparrow$ & IoU$\uparrow$ & CD$\downarrow$ & F-Score$\uparrow$ & IoU$\uparrow$ \\
\midrule
Baseline ($\mathcal{Q}_\text{learnable}$) & \secondcell{0.0072} & \bestcell{0.9300} & \thirdcell{0.7603} & \secondcell{0.0318} & \thirdcell{0.8127} & \secondcell{0.6912} \\
\midrule
w/o DINOv3 features & 0.0077 & 0.9282 & 0.7552 & \thirdcell{0.0324} & 0.8024 & 0.6715 \\
w/o Utonia features & 0.0080 & 0.9268 & 0.7477 & 0.0492 & 0.7671 & 0.6478 \\
w/o diversity loss & \thirdcell{0.0073} & \thirdcell{0.9286} & 0.7551 & 0.0353 & 0.8031 & 0.6695 \\
w/o multi-hot & 0.0078 & 0.9214 & 0.7469 & 0.0394 & 0.8015 & 0.6568 \\
\midrule
w/ $\mathcal{Q}_\text{mask}$ & \bestcell{0.0067} & \secondcell{0.9289} & \secondcell{0.7626} & 0.0336 & \secondcell{0.8151} & \thirdcell{0.6880} \\
w/ $\mathcal{Q}_\text{point}$ & 0.0075 & 0.9274 & \bestcell{0.7655} & \bestcell{0.0270} & \bestcell{0.8216} & \bestcell{0.7051} \\
\bottomrule
\end{tabular}}
\end{minipage}
\vspace{-1 em}
\end{table*}

\Cref{tab:depth_ablation_combined} (a) shows that \method preserves strong visible-region geometry after latent decomposition.
Among 3D reconstruction baselines, \method reduces AbsRel by 14.9\% and RMSE by 10.3\% over the best baseline, while also achieving the highest completion score.
Compared with the MoGe2 reference, \method produces competitive visible-region depth while additionally yielding decomposable meshes.
Qualitative depth comparisons are provided in~\cref{fig:vis_depth}.

\subsection{Ablation Study}
\label{sec:exp_ablation}

We ablate \method along two axes: decomposition design choices and object-query type.
All variants are trained solely on the 3D-FRONT InstPIFU train split using a single H100 GPU, with all other training hyperparameters held identical.
The learnable-query variant serves as the prompt-free baseline, and each row in~\cref{tab:depth_ablation_combined} (b) changes one component or query modality.

\noindent\textbf{Importance of Feature Augmentation}
Feature augmentation is important because latent decomposition must distinguish object extent inside a shared scene representation, not only preserve the global scene shape.
Removing DINOv3 features drops object IoU by $0.0197$, while removing Utonia increases object CD by $54.7\%$ over the learnable-query baseline.

\noindent\textbf{Ablations on Loss Design}
The multi-hot mask supervision and query diversity loss are significant for improving decomposition quality.
We investigate them by removing the diversity loss and replacing multi-hot supervision with single-hot labels.
The former variant drops object IoU by $3.1\%$,
while the latter reduces object IoU by $5.0\%$.

\noindent\textbf{Ablations on Query Type}
Our query-based decomposition design is flexible to various query types.
The mask variant achieves the best scene-level performance (scene CD reduced by $6.9\%$),
while the point variant achieves the best object-level performance (object CD reduced by $15.1\%$ over learnable queries),
suggesting that explicit prompts further improve decomposition quality.

\section{Conclusions}\label{sec:conclusions}

We present \method,
a novel paradigm for Composable Latent 3D decomposition.
Our approach departs from the common recipe of segmenting objects in 2D and reconstructing each one separately, by performing decomposition directly in a \emph{whole}-scene latent space.
It achieves state-of-the-art results in both object quality and scene-level consistency across various datasets,
in a significantly faster manner.
Our simple, flexible, and scalable query-based transformer also supports various query types and latent sources.
It also achieves visible-region depth accuracy on par with strong depth estimators such as MoGe2,
which is another benefit over per-object generation methods.

\textbf{Limitations and future work.}
The coarse latent grid can absorb thin or tightly clustered objects into one cell; a hierarchical latent is a natural remedy.
Training data is also a bottleneck: available scene datasets are small in scale, limited in quality, and sparsely populated with objects. Scaling to larger, higher-quality, and more densely populated data is an important next step.

\begin{ack}
We thank Ruijie Zhu for insightful discussions that helped refine this work, and Yingqi Wang for valuable assistance with the robotic demonstration.
Yongcai Wang is supported in part by the National Natural Science Foundation of China under Grant No.~61972404, the Public Computing Cloud, Renmin University of China, and the Blockchain Lab, School of Information, Renmin University of China.
Deying Li is supported in part by the National Natural Science Foundation of China under Grant No.~12071478.
Chuanxia Zheng is supported by NTU SUG-NAP and National Research Foundation, Singapore, under its NRF Fellowship Award NRF-NRFF17-2025-0009.
\end{ack}

\appendix
\clearpage

\begin{center}
    {\Large\textbf{\method: Composable Latent 3D Decomposition}}

    \vspace{0.1in}
    {{\textit{Supplementary Material}}}

\end{center}

\section{Additional Method Details}
\label{sec:appendix_experiment}

\subsection{Attention Block Details}
\label{sec:appendix_attention}

\begin{figure}[H]
    \centering
    \includegraphics[width=0.62\linewidth]{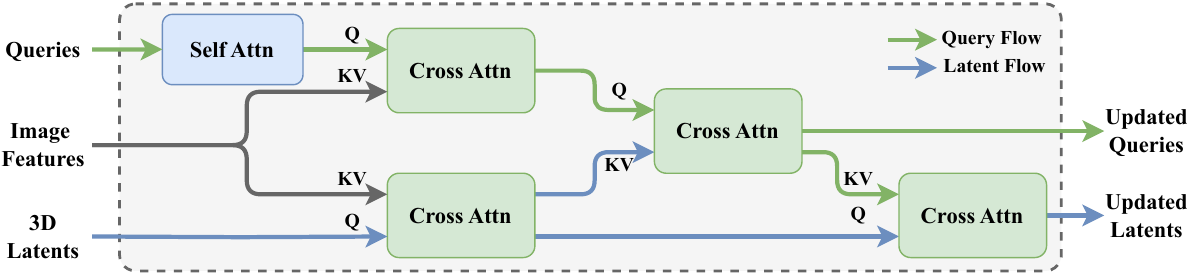}
    \caption{\textbf{Architecture details of the attention block used in the Query-Based 3D Latent Decomposition Transformer.}}
    \label{fig:supp_attn}
\end{figure}

This section expands the \emph{Attention Blocks} component of~\cref{sec:ladm}.
As shown in~\cref{fig:supp_attn}, each block maintains two information streams:
object queries, which represent instance hypotheses, and 3D scene-latent tokens, which preserve the geometric support of the reconstructed scene.
The block first applies self-attention over the query tokens so that different object hypotheses can exchange context before being grounded in visual evidence.
The queries and latent tokens then both cross-attend to the frozen DINOv3 image features, using image features as keys and values.
This step injects 2D semantic cues into both streams while keeping the 3D latent tokens anchored to the shared scene coordinate frame.

After image-feature grounding, the block exchanges information between the two streams through bidirectional query--latent attention.
The query stream first attends to the latent-conditioned evidence, allowing each query to refine the spatial extent of its object hypothesis.
The latent stream then attends back to the updated queries, making each 3D latent token more instance-aware before mask prediction.
Stacking these blocks gradually turns generic queries into object-specific tokens and enriches the scene latent with instance-level cues, which is the representation used by the final assignment head in~\cref{sec:ladm}.

\heading{Positional Encoding}
We use 3D rotary position embeddings (RoPE)~\cite{su2024roformer} in the query self-attention and bidirectional query--latent cross-attention layers.
For an attention query/key feature $\bm{x}\in\mathbb{R}^{D_\text{head}}$ with 3D position $(p_x,p_y,p_z)$, we partition its 2D feature pairs across the three spatial axes.
Let $\bm{x}_\alpha$ denote the feature pairs assigned to axis $\alpha\in\{x,y,z\}$, and let $P_\alpha$ be the number of pairs assigned to that axis.
For each axis, the $i$-th 2D feature pair in $\bm{x}_\alpha$ is rotated by
\begin{equation}
    \theta_{i,p_\alpha} = \frac{p_\alpha}{\lambda^{i/P_\alpha}},
    \quad i\in\{0,\ldots,P_\alpha-1\},
\end{equation}
where $\lambda$ is the RoPE frequency base.
The rotation is applied independently for each axis,
\begin{equation}
    \begin{pmatrix}
        x'_{\alpha,2i} \\
        x'_{\alpha,2i+1}
    \end{pmatrix}
    =
    \begin{pmatrix}
        \cos\theta_{i,p_\alpha} & -\sin\theta_{i,p_\alpha} \\
        \sin\theta_{i,p_\alpha} & \cos\theta_{i,p_\alpha}
    \end{pmatrix}
    \begin{pmatrix}
        x_{\alpha,2i} \\
        x_{\alpha,2i+1}
    \end{pmatrix}.
\end{equation}
The final head feature concatenates the axis-wise rotated pairs, allowing attention to encode relative displacement along all three spatial axes.
In our implementation, the attention head dimension is $D_\text{head}=64$, which gives $D_\text{head}/2=32$ complex feature pairs.
We allocate 10 frequency pairs to each of the $x$, $y$, and $z$ axes and use identity-phase padding for the remaining two pairs.

The 3D latent tokens use their integer sparse-grid coordinates as RoPE positions.
For query tokens, however, there is no native 3D coordinate because they are 1D object-hypothesis tokens.
Following the M-RoPE design in Qwen2-VL~\cite{wang2024qwen2vl}, we lift the query index to a virtual diagonal 3D coordinate, \ie query $k$ is assigned $(k,k,k)$, such as $(0,0,0)$, $(1,1,1)$, and $(2,2,2)$.
This simple diagonal extension gives query--query and query--latent attention a consistent 3D positional frame, while the latent stream still uses the actual scene geometry coordinates.

\subsection{Training Data Processing}
\label{sec:appendix_dataset}
We derive multi-hot part supervision from the multi-object O-Voxel representation before encoding.
Each object is represented by its active O-Voxel cells with a contiguous instance index, and cells from all objects are merged by high-resolution spatial coordinates instead of being forced into a single object label.
When multiple objects share a coordinate, their instance labels are accumulated into an object-channel multi-hot vector, while geometric attributes are merged by averaging vertex offsets and OR-aggregating intersection indicators; single-object coordinates naturally reduce to one-hot labels.
Because mask prediction is performed after the O-Voxel encoder, we project these full-resolution multi-hot labels to the latent sparse grid using the encoder downsampling ratio, which is $16\times$ for the $1024\!\rightarrow\!64$ grid compression in our implementation, and use channel-wise OR to aggregate labels within each latent cell.
The resulting target allows one active latent token to supervise multiple object channels at contact or overlapping regions; the predicted per-query sigmoid masks are then matched to these object masks and trained with multi-label BCE and per-object Dice losses as described in~\cref{sec:training}.

\subsection{Evaluation Data Processing}
\label{sec:appendix_eval_data}
We construct evaluation sets from both synthetic datasets, 3D-FRONT~\cite{fu20213d} and MetaScenes~\cite{yu2025metascenes}, and real-world scan datasets, Replica~\cite{replica19arxiv} and ScanNet++ v2~\cite{yeshwanth23scannet:}.
For the main benchmarks, we retain scenes or views with 2--8 visible objects and exclude samples with severe object collisions, unreliable segmentations, or low-quality ground-truth meshes, so that the metrics reflect reconstruction quality rather than invalid evaluation geometry.
For MetaScenes-hard, the separate 50-scene test set used for the scalability evaluation in~\cref{sec:exp_scalability}, we retain scenes with 9--20 visible objects while applying the other filtering criteria unchanged.

\heading{3D-FRONT}
3D-FRONT~\cite{fu20213d} is a large-scale synthetic indoor-scene dataset of professionally designed furnished rooms, with semantic layouts and textured 3D furniture models.
Its complete object meshes and layouts make it suitable for evaluating scene-level geometry and object-level decomposition.
We start from the 3D-FRONT test split provided by InstPIFU~\cite{liu2022instpifu} and apply the common filtering protocol above, resulting in 85 high-quality scenes.

\heading{MetaScenes}
MetaScenes~\cite{yu2025metascenes} is a large-scale, simulatable 3D scene dataset built for automated replica creation from real-world indoor scans.
It combines scan-derived layouts with clean object assets and physically plausible scene optimization, providing synthetic scenes with more realistic arrangements.
We randomly sample camera views from the full scene set, render input images, filter the rendered samples with the same protocol as 3D-FRONT, and randomly select 50 high-quality scenes for evaluation.

\heading{Replica}
Replica~\cite{replica19arxiv} contains photorealistic indoor scene reconstructions with dense geometry, high-resolution textures, and semantic/instance annotations.
We use the processed and rendered Replica data released by DP-Recon~\cite{ni2025dprecon}.
After filtering for reliable scan geometry, segmentation quality, and the common object-count range, the Replica evaluation set contains 8 scenes and 14 images.

\heading{ScanNet++ v2}
ScanNet++~\cite{yeshwanth23scannet:} is a high-fidelity indoor-scene dataset with accurate scan geometry, high-resolution imagery, RGB-D captures, and semantic annotations.
Its complex real-world scenes provide a more challenging setting for assessing model performance under realistic scan conditions.
Applying the same real-scan filtering protocol yields 45 scenes and 56 images.

\subsection{Training Details}
\label{sec:appendix_training}

\heading{TRELLIS.2 Scene-Level Finetuning}
The original TRELLIS.2~\cite{xiang2025native} is trained on a large-scale dataset of 3D assets, but with a focus on \emph{single-object} generation.
To adapt it for scene-level meshes $\mathcal{M}^s$, we finetune its three geometry components---the sparse-structure Flow Transformer, the O-Voxel Flow Transformer, and the O-Voxel decoder---on our curated \emph{scene}-level datasets, following the same data processing and training losses as TRELLIS.2.
We use the 1024-resolution SLAT DiT setting for this scene-level finetuning.
We find that TRELLIS.2 can already produce promising scene generation after finetuning on a relatively small-scale 3D scene dataset, demonstrating the strong geometry generalizability of the foundation model.

\heading{Dice Loss Formulation}
The Dice term used in the mask loss $\mathcal{L}_{\mathrm{mask}}$ is computed per matched query as
\begin{equation}
    \mathrm{Dice}\!\left(\mathbf{S}_{:,\pi(k)},\,\bm{y}^{*}_{k}\right)
    =1-\frac{2\,\langle\mathbf{S}_{:,\pi(k)},\,\bm{y}^{*}_{k}\rangle+\epsilon}
    {\|\mathbf{S}_{:,\pi(k)}\|_{1}+\|\bm{y}^{*}_{k}\|_{1}+\epsilon},
\end{equation}
where $\langle\cdot,\cdot\rangle$ is the Euclidean inner product,
$\|\cdot\|_{1}$ the $\ell_{1}$ norm,
and $\epsilon$ a small constant for numerical stability in degenerate zero-norm cases.
Computing Dice \emph{per query} (macro averaging) gives small and large objects equal weight,
which is essential in our sparse latent space where a per-cell BCE alone is dominated by large objects that contain many more active latent cells.

\subsection{Inference Time Details}
\label{sec:appendix_timing}

\begin{table}[H]
\centering
\caption{\textbf{Detailed per-dataset inference time.}
We report the average per-scene inference time in seconds.}
\label{tab:supp_timing}
\small
\setlength{\tabcolsep}{3pt}
\resizebox{0.6\linewidth}{!}{
\begin{tabular}{@{}l c cccc c@{}}
\toprule
\textbf{Method} & \textbf{Prior} & \textbf{3D-FRONT}~\cite{fu20213d} & \textbf{MetaScenes}~\cite{yu2025metascenes} & \textbf{Replica}~\cite{replica19arxiv} & \textbf{ScanNet++}~\cite{yeshwanth23scannet:} & \textbf{Avg.} \\
\midrule
Gen3DSR~\cite{gen3dsr}        & \priormask  & 215.90 & 242.90 & 282.60 & 264.80 & 251.55 \\
PartCrafter~\cite{lin2025partcrafter} & \priornone  & 49.17 & 52.91 & 62.79 & 56.34 & 55.30 \\
MIDI~\cite{huang2025midi}  & \priormask  & 64.20 & 70.70 & 82.50 & 75.20 & 73.15 \\
SceneGen~\cite{meng2026scenegen} & \priormask  & 48.60 & 50.00 & 59.70 & 62.50 & 55.20 \\
SAM 3D~\cite{chen2026sam}     & \priormask  & 30.90 & 33.40 & 39.10 & 35.00 & 34.60 \\
3D-Fixer~\cite{threedfix}     & \priormask  & 93.20 & 103.50 & 120.90 & 93.90 & 102.88 \\
Point-SAM~\cite{zhou2025pointsam} & \priorpoint & 20.18 & \thirdcell{13.55} & 23.91 & 19.22 & 19.22 \\
\midrule
\textbf{\method (Free)}  & \priornone  & \secondcell{19.87} & \secondcell{13.40} & \secondcell{23.43} & \secondcell{18.85} & \secondcell{18.89} \\
\textbf{\method (Point)} & \priorpoint & \thirdcell{20.12} & \bestcell{13.08} & \thirdcell{23.76} & \thirdcell{19.17} & \thirdcell{19.03} \\
\textbf{\method (Mask)}  & \priormask  & \bestcell{19.55} & 13.71 & \bestcell{23.19} & \bestcell{18.47} & \bestcell{18.73} \\
\bottomrule
\end{tabular}
}
\end{table}

\cref{tab:supp_timing} provides the per-dataset timing details behind the averaged Time columns in the main quantitative tables.
All timings are measured on one NVIDIA L40S 48 GB GPU with 8 vCPUs (AMD EPYC 7R13) and 64 GB RAM.
We exclude texture-related inference and processing time from all timing reports.

\subsection{Inference Details}
\label{sec:appendix_inference}

For Learnable queries, the initial multi-hot assignment may still contain redundant queries or spatially incoherent predictions,
so we apply a lightweight two-step refinement at inference time.
We first merge redundant queries that describe the same object via greedy non-maximum suppression,
fusing any pair whose mask intersection-over-union exceeds $0.5$.
We then extract the 6-connected components of each instance over the sparse latent grid and reassign small fragments to neighboring instances by majority vote,
removing isolated noise while preserving compact small objects.

Using the refined discrete mask,
we carve the per-instance latent out of the scene latent by gathering the cells assigned to each query,
\begin{equation}
    \bm{z}^o_k=\{(\hat{\bm{f}}_i,\,\hat{\bm{p}}_i) \mid m_{i,k}=1\},
\end{equation}
and decode each $\bm{z}^o_k$ in parallel with the shared TRELLIS.2 decoder $\mathcal{D}$ to obtain the per-object meshes $\{\hat{\mathcal{M}}^o_k\}_{k=1}^{\hat{K}}$,
where $\hat{K}$ is the predicted number of objects.
Since all instances are carved from a single scene latent in a shared coordinate system,
the decoded meshes are naturally aligned and faithfully preserve the global spatial relationships of the original scene,
without any post-hoc registration or layout regression as required by other methods~\cite{huang2025midi,meng2026scenegen,chen2026sam}.

\section{Additional Experimental Results}
\label{sec:appendix_results}

\subsection{Geometric Interpenetration}
\label{sec:appendix_interpenetration}

We quantify object interpenetration on 3D-FRONT and MetaScenes, where complete ground-truth object meshes are available.
We report two complementary metrics on the reconstructed scenes.
\emph{Collision rate} measures the percentage of object pairs with non-zero mesh penetration.
\emph{Normalized interpenetration volume} measures the pairwise intersection volume normalized by object volume, computed by voxelizing each scene on a $256^3$ grid.

\begin{table}[htbp]
\centering
\caption{\textbf{Geometric interpenetration on synthetic datasets.}
We report collision rate (CR) and normalized interpenetration volume (NIV), both in percent; lower is better.
\textbf{Prior} column: \protect\priormask{} denotes a 2D instance mask, \protect\priorpoint{} a 3D point prompt, and \protect\priornone{} no prior.
We highlight the \besthl{Best}, \secondhl{Second-Best}, and \thirdhl{Third-Best} results.}
\label{tab:collision_results}
\small
\setlength{\tabcolsep}{9pt}
\begin{tabular}{@{}l c cc cc@{}}
\toprule
\multirow{2}{*}{\textbf{Method}} & \multirow{2}{*}{\textbf{Prior}} & \multicolumn{2}{c}{\textbf{3D-FRONT}} & \multicolumn{2}{c}{\textbf{MetaScenes}} \\
\cmidrule(lr){3-4} \cmidrule(lr){5-6}
& & CR$\downarrow$ & NIV$\downarrow$ & CR$\downarrow$ & NIV$\downarrow$ \\
\midrule
Gen3DSR~\cite{gen3dsr} & \priormask & \secondcell{1.9} & \thirdcell{0.12} & \bestcell{1.1} & \bestcell{0.03} \\
MIDI~\cite{huang2025midi} & \priormask & 8.8 & 0.34 & 5.9 & 0.32 \\
SceneGen~\cite{meng2026scenegen} & \priormask & 5.1 & 0.35 & 17.6 & 1.11 \\
SAM 3D~\cite{chen2026sam} & \priormask & 3.6 & \thirdcell{0.12} & 10.1 & 0.91 \\
PartCrafter~\cite{lin2025partcrafter} & \priornone & 2.8 & 1.18 & \thirdcell{2.8} & \secondcell{0.12} \\
\midrule
\textbf{\method (Free)} & \priornone & \bestcell{1.8} & \bestcell{0.06} & \secondcell{2.3} & \thirdcell{0.13} \\
\textbf{\method (Point)} & \priorpoint & 3.4 & 0.19 & 4.5 & 0.19 \\
\textbf{\method (Mask)} & \priormask & \thirdcell{2.4} & \secondcell{0.11} & 4.3 & 0.22 \\
\bottomrule
\end{tabular}
\end{table}

\Cref{tab:collision_results} shows that \method (Free) achieves the lowest collision rate and normalized interpenetration volume on 3D-FRONT, at 1.8\% and 0.06\%, respectively.
On MetaScenes, it ranks second in collision rate at 2.3\% and third in normalized interpenetration volume at 0.13\%.
All three \method variants achieve lower values on both metrics than MIDI and SceneGen across the two datasets.

\subsection{Decomposition from Existing Meshes}
\label{sec:appendix_mesh_decompose}

\begin{figure}[H]
    \centering
    \includegraphics[width=0.8\linewidth]{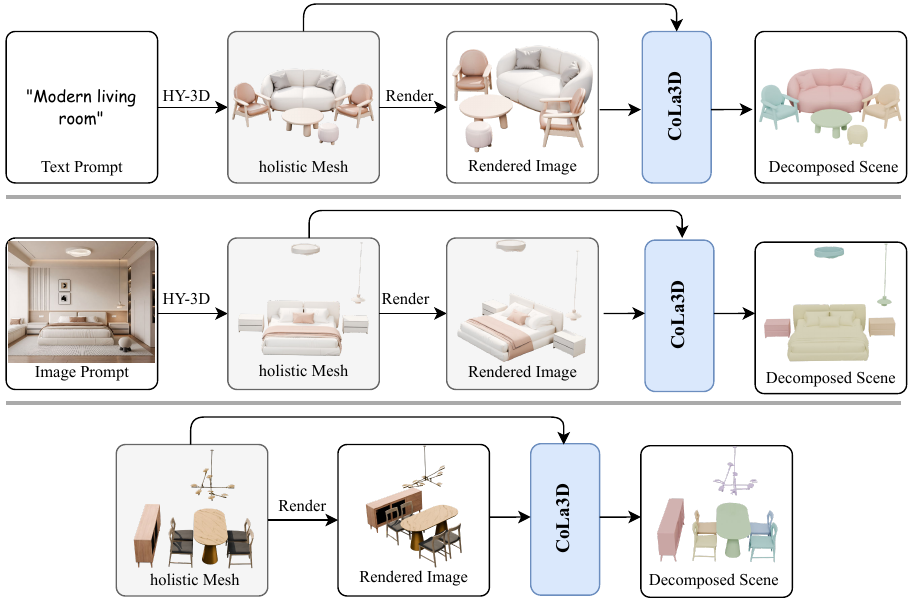}
    \caption{\textbf{Decomposition from holistic scene meshes.} \method produces coherent object-level meshes across generated and existing holistic scene meshes.}
    \label{fig:mesh_decompose}
\end{figure}

Beyond image input, the same query transformer also accepts encoded scene meshes.
\cref{fig:mesh_decompose} shows three mesh-decomposition pathways: two HunYuan-3D 3.1~\cite{hunyuan3d31website} meshes generated from text and image prompts, respectively, and an existing holistic scene mesh.
In each case, \method consumes the holistic mesh, renders a proxy image, and decomposes the encoded scene latent into aligned object meshes.
The results show that the same mesh-to-mesh decomposition pipeline applies cleanly to generated and existing scene meshes.

\subsection{Additional Synthetic Visualizations}
\label{sec:appendix_vis_synthetic}

We provide more qualitative comparisons on the 3D-FRONT InstPIFU split and MetaScenes, using the same layout as the main-paper synthetic visualization in~\cref{fig:vis_synthetic}.
These examples further show how \method preserves object alignment and relative scale across diverse synthetic indoor scenes.

\begin{figure}[H]
    \centering
    \resizebox{\linewidth}{!}{\includegraphics{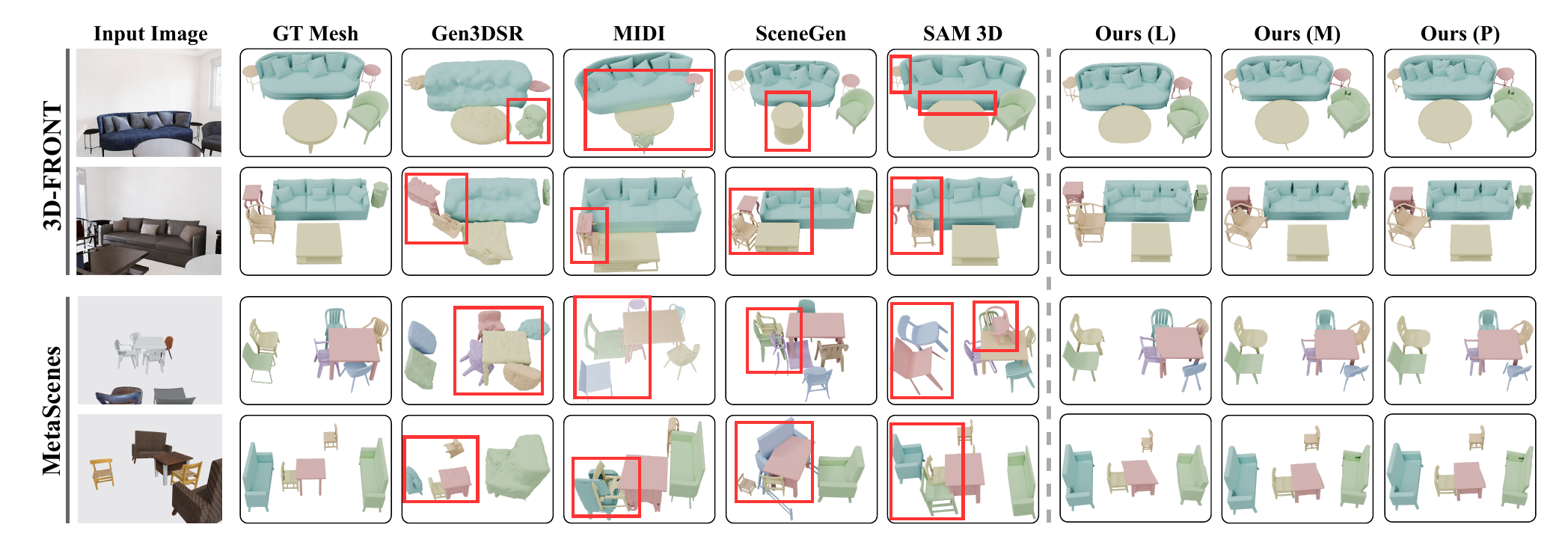}}
    \caption{\textbf{Additional qualitative comparisons on synthetic scenes.} We show more results on the 3D-FRONT InstPIFU split and MetaScenes following the same layout as~\cref{fig:vis_synthetic}.}
    \label{fig:supp_vis_synthetic}
\end{figure}

\subsection{Additional Real-World Visualizations}
\label{sec:appendix_vis_real}

We provide more zero-shot qualitative comparisons on Replica and ScanNet++, using the same layout as the main-paper real-world visualization in~\cref{fig:vis_real}.
These examples illustrate the behavior of \method under real image observations with partial scans and cluttered layouts.

\begin{figure}[H]
    \centering
    \includegraphics[width=\linewidth]{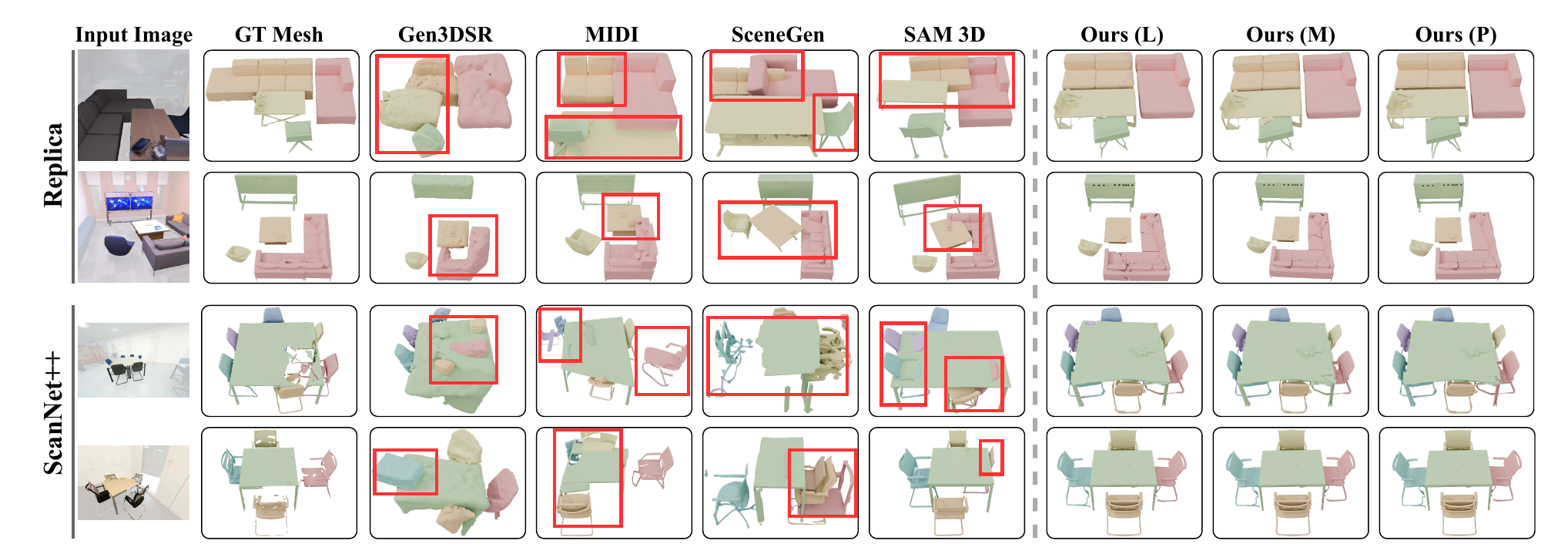}
    \caption{\textbf{Additional qualitative comparisons on real-world scenes.} We show more zero-shot results on Replica and ScanNet++ following the same layout as~\cref{fig:vis_real}.}
    \label{fig:supp_vis_real}
\end{figure}

\subsection{Qualitative Depth Comparison}
\label{sec:appendix_vis_depth}

We provide qualitative depth visualizations on 3D-FRONT to complement the quantitative results in~\cref{tab:depth_ablation_combined}.

\begin{figure}[H]
    \centering
    \includegraphics[width=\linewidth]{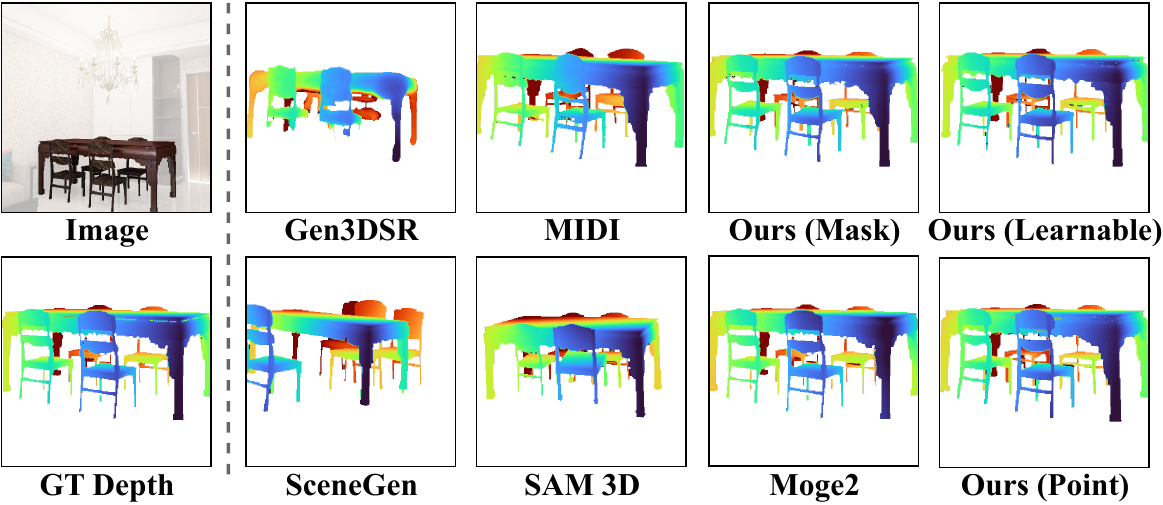}
    \caption{\textbf{Qualitative depth comparison on 3D-FRONT.} We visualize the input image, ground-truth depth, rendered depth maps from Gen3DSR, MIDI, SceneGen, SAM 3D, and \method with mask, learnable, and point queries, and the MoGe2 prediction as a monocular-depth reference. \method yields visible-region depth on par with MoGe2 while additionally producing decomposable per-object geometry.}
    \label{fig:vis_depth}
\end{figure}

\section{Broader Impacts}
\label{sec:appendix_broader_impacts}

\method turns a single image or a holistic scene mesh into an editable, object-level 3D scene in a single forward pass.
On the positive side, this lowers the barrier to producing simulation-ready 3D environments and can benefit downstream applications such as robotics and embodied-AI training, physically plausible simulation, AR/VR experiences, digital twins, and 3D content creation for games, film, and design.
By decomposing scenes directly in a shared latent space, our method also removes the need for per-object reconstruction and post-hoc layout regression, which makes such tools more accessible to researchers and practitioners with limited compute.

On the negative side, stronger composable 3D reconstruction tools may be misused to fabricate realistic but fake environments, or to reconstruct private interior spaces from casually captured images without consent, raising potential privacy, copyright, and disinformation concerns.
The model also inherits biases from its training data: scene datasets are dominated by certain categories of indoor furniture and room layouts, so generation and decomposition quality may be uneven across cultures, building styles, and underrepresented object categories.
We encourage downstream users to respect dataset licenses, obtain consent before reconstructing private spaces, and follow applicable privacy and safety requirements when integrating such 3D reconstruction tools into user-facing products.

\end{document}